# MPEC: Manifold-Preserved EEG Classification via an Ensemble of Clustering-Based Classifiers

Shermin Shahbazi
*Department of Electrical and Computer Engineering,*
*University of Zanjan,*
Zanjan, Iran.
sh.shahbazi@znu.ac.ir
ORCID: 0009-0008-5402-1313

Mohammad-Reza Nasiri
*Department of Computer Science and Information Technology,*
*Institute for Advanced Studies in Basic Sciences (IASBS)*
Zanjan 45137-66731, Iran.
mr.nasiri@iasbs.ac.ir
ORCID: 0009-0002-6962-6311

Majid Ramezani
*Department of Computer Science and Information Technology,*
*Institute for Advanced Studies in Basic Sciences (IASBS)*
Zanjan 45137-66731, Iran.
ramezani@iasbs.ac.ir
ORCID: 0000-0003-0886-7023

*Abstract*— Accurate classification of EEG signals is crucial for brain-computer interfaces (BCIs) and neuroprosthetic applications, yet many existing methods fail to account for the non-Euclidean, manifold structure of EEG data, resulting in suboptimal performance. Preserving this manifold information is essential to capture the true geometry of EEG signals, but traditional classification techniques largely overlook this need. To this end, we propose MPEC (Manifold-Preserved EEG Classification via an Ensemble of Clustering-Based Classifiers), that introduces two key innovations: (1) a feature engineering phase that combines covariance matrices and Radial Basis Function (RBF) kernels to capture both linear and non-linear relationships among EEG channels, and (2) a clustering phase that employs a modified K-means algorithm tailored for the Riemannian manifold space, ensuring local geometric sensitivity. Ensembling multiple clustering-based classifiers, MPEC achieves superior results, validated by significant improvements on the BCI Competition IV dataset 2a.

## I. INTRODUCTION

EEG signal classification is essential in brain-computer interfaces (BCIs) and neuroprosthetics, where precise interpretation supports real-time control and cognitive applications. However, traditional techniques often overlook the non-Euclidean, manifold structure of EEG data, leading to suboptimal results [1]. We propose *Manifold-Preserved EEG Classification via an Ensemble of Clustering-Based Classifiers (MPEC)*, a novel method that enhances classification accuracy by preserving the intrinsic manifold structure of EEG signals. By leveraging the Riemannian manifold context, MPEC captures and utilizes the geometric relationships within the data for improved classification performance.

Preserving manifold information is crucial due to the non-linear, high-dimensional nature of EEG signals, which reside in a curved, non-Euclidean space [2]. Traditional Euclidean-based methods often distort signal relationships, leading to misclassification [3]. By maintaining the manifold structure, our approach accurately captures data geometry and preserves local relationships, essential for applications in BCIs, neuroprosthetics, and diagnostics, where precise EEG interpretation significantly impacts performance. Most EEG classification systems ignore the underlying manifold structure, treating signals in a linear, Euclidean space [1][2]. This simplification restricts their capacity to model complex, non-linear dependencies between EEG channels, reducing classification accuracy. A notable gap exists in the literature for methods that incorporate the manifold context of EEG data in feature engineering and classification.

To fill this gap, we introduce a solution that preserves the EEG signal manifold structure throughout the classification pipeline. Our approach starts with *feature engineering* using covariance matrices and Radial Basis Function (RBF) kernels to capture both linear and non-linear relationships. Then a novel *clustering* algorithm, tailored for Riemannian space, leverages a custom metric based on Riemannian distance and curvature. Finally, *dimensionality reduction* projects clusters onto the tangent space, and an *ensemble of several weak classifiers*, combined via stacking, boosts accuracy while maintaining the manifold context.

Our method improves EEG classification by clustering data in manifold space and using a distance metric that captures both global and local geometries, ensuring sensitive classification. The ensemble model enhances accuracy and robustness, outperforming traditional Euclidean methods. Our study is driven by the following key research questions:

**RQ1**: Does preserving the manifold structure of EEG signals significantly improve classification accuracy, and to what extent does this enhancement manifest in various classification scenarios?

**RQ2**: How does the application of ensemble classification methods on locally clustered Riemannian manifolds contribute to the performance improvement of EEG signal classification compared to traditional approaches?

**RQ3**: What advantages does the novel K-means algorithm utilizing a distance metric that combines Riemannian distance and curvature offer for clustering high-dimensional EEG data?

**RQ4**: Does projecting data points from each Riemannian cluster onto a lower-dimensional representation maintain sufficient discriminative power necessary for accurate EEG classification?

The paper is structured as follows: Section 2 reviews EEG classification literature, highlighting methods that overlook manifold structure. Section 3 details our methodology, including feature engineering, clustering, and ensemble learning. In Section 4, we present results, comparing our method to traditional approaches. Section 5 concludes with implications and future research directions.

## II. LITERATURE REVIEW

EEG classification is complex due to EEG signals' non-linear, high-dimensional nature in a non-Euclidean space. Recent research focuses on dimensionality reduction, geometric techniques, and ensemble learning to address this. Dimensionality reduction is essential in EEG classification to handle high-dimensional data and improve efficiency. The authors in [4] propose a novel method for Symmetric Positive Definite (SPD) data, mapping it to tangent spaces to retain geometry, while [5] introduces geometry-aware techniques for SPD manifolds, balancing classification power and computational cost.

Riemannian geometry-based approaches have gained attention in EEG classification due to the data's non-linearity and manifold structure. A review in [2] highlights the advantages of Riemannian methods in handling EEG's noise sensitivity and non-stationarity, particularly in BCIs. Similarly, [6] shows that Riemannian kernel methods enhance BCI classification performance compared to traditional filtering. Additionally, [7] introduces a manifold learning-based ensemble method using Riemannian geometry for higher accuracy. These studies emphasize the importance of preserving EEG's non-Euclidean structure for improved classification.

Ensemble learning enhances EEG classification by combining multiple weak classifiers into a robust model. In [8], stacking ensembles for emotion classification combine Random Forest, LightGBM, and Gradient Boosting with a Random Forest meta-model. Similarly, [7] proposes an ensemble method called MLCSP-TSE-MLP, using tangent space embedding to reduce computational costs and improve performance on large EEG datasets. Stacking ensemble learning effectively addresses the variability within EEG data, benefiting complex tasks in BCIs and emotion recognition.

EEG classification research has focused on application-specific methods. For instance, [9] combines CNNs with continuous wavelet transforms for epileptic seizure classification, using time-frequency images to capture seizure patterns. In motor imagery, [10] employs a hybrid approach with Kernel Extreme Learning Machines, PCA, and Fisher's Linear Discriminant for high accuracy in BCI tasks.

Recent research has introduced innovations to address EEG classification's computational and structural challenges. In [1], the authors review EEG classification algorithms, highlighting adaptive classifiers and tensor-based approaches suited for EEG's high-dimensional nature. While deep learning techniques like convolutional neural networks are gaining popularity, Riemannian geometry methods continue to be effective for handling EEG's manifold structure. Additionally, [11] offers the EEGLAB toolbox, integrating independent component analysis and time-frequency decompositions, which aids researchers in managing complex EEG datasets. These advancements enhance both accuracy and accessibility in EEG signal processing.

## III. METHODS

### A. Dataset

The *BCI Competition IV Dataset 2a* [12] is used in this study, containing EEG recordings from 9 healthy subjects performing motor imagery tasks across two sessions. The data includes 25 channels (22 EEG and 3 EOG) sampled at 250 Hz, with four motor imagery tasks: *left hand*, *right hand*, *both feet*, and *tongue*. Each session has 288 trials, totaling 576 trials per subject.

### B. Proposed EEG Signal Classification System

This study presents *MPEC: Manifold-Preserved EEG Classification via an Ensemble of Clustering-Based Classifiers*, a method that preserves the geometric structure of the Riemannian manifold by retaining curvature-related information in EEG signals, ensuring essential non-Euclidean relationships are maintained for accurate classification. As depicted in Fig. 1 MPEC consists of four phases: *feature engineering*, *data clustering*, *dimensionality reduction*, and *classification* detailed in the following sections.

#### 1) Phase 1: Feature Engineering

In this phase, we use a three-step approach: (A) selecting relevant EEG channels based on correlation with target labels, (B) applying the covariance matrix to capture linear relationships between channels, and (C) using RBF kernels to identify non-linear relationships among the channels. The rationale behind starting with feature selection is to focus on the most informative channels, enhancing classification performance by filtering out less relevant signals. The covariance matrix then allows us to analyze how channels co-vary, revealing fundamental patterns in brain activity. In parallel, RBF kernels capture non-linear interactions that may otherwise go unnoticed, offering a more comprehensive representation of the data. This combination effectively leverages both linear and non-linear relationships in the EEG signals.

**A. Feature Selection:** in the first step of feature extraction, we perform feature selection on the raw EEG signals to identify the most relevant channels for classification. We calculate the correlation between each channel's signal and the target labels, considering channels with higher correlation values as more informative. This correlation-based method reduces the input data's dimensionality by removing less relevant features, improving the model's efficiency and accuracy. By focusing on key channels, we preserve crucial information, minimize noise, and reduce computational load, resulting in a more robust and interpretable model for the next stages.

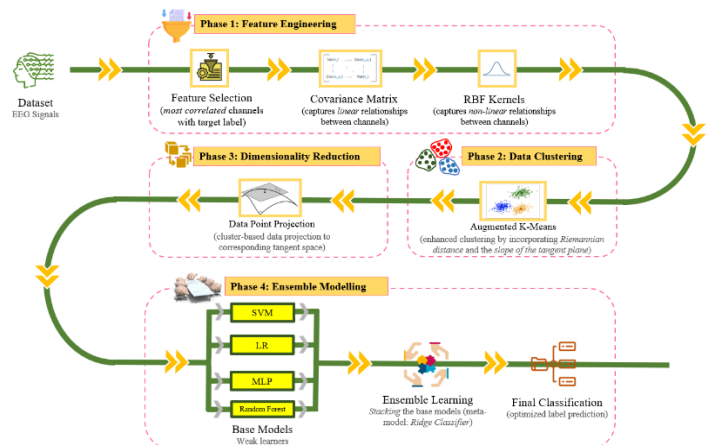

Fig. 1. The general architecture of the proposed method (MPEC)

**B. Covariance Matrices**: covariance matrices effectively capture statistical relationships between EEG channels in motor imagery tasks, summarizing dependencies across electrodes for tasks like imagined hand movements [13]. This representation encapsulates critical brain activity patterns and reduces raw EEG data dimensionality [13]. However, the non-Euclidean nature of covariance matrices challenges traditional Euclidean methods, which can distort their geometry [14]. To address this, we process covariance matrices within the Riemannian manifold space, preserving their inherent curvature. This approach enables more accurate distance computations between matrices, reflecting physiological differences in brain activity across motor imagery tasks. In the Riemannian manifold, we calculate geodesic distances, which represent the shortest paths on the manifold and preserve true geometric relationships between signals [14]. This method enhances classification by offering discriminative features, reduces noise sensitivity, and improves robustness and accuracy in EEG-based BCI systems.

In our feature engineering, we use covariance matrices to capture relationships between EEG channels during motor imagery tasks. Given a dataset $X = \{x_1, x_2, \ldots, x_N\}$, where each $x_i \epsilon \mathbb{R}^n$ represents a sample of the EEG signal, the covariance matrix $C \in \mathbb{R}^{n \times n}$ is computed as:

$$C = \frac{1}{N-1}\sum_{k=1}^{N}(x_i - \mu)(x_i - \mu)^T \quad (1)$$

here, $\mu$ is the mean vector, $x_i$ is a sample, and $N$ is the total number of samples. The covariance matrix captures dependencies between channels, providing a detailed statistical representation of the signal.

Covariance matrices need to be *symmetric* and *positive definite* to belong to the *Symmetric Positive Definite (SPD)* space, which is crucial for performing meaningful geometric operations like distance calculations in the Riemannian framework [15]. The proof of this is provided below.

**Symmetry:** the covariance matrix is inherently symmetric, as shown in (1), where $\mu$ represents the dataset mean and $(x_i - \mu)$ the deviation of each sample from the mean. Since covariance between two variables is symmetric ($C_{ij} = C_{ji}$ for all $i$ and $j$), this matrix symmetry reflects the bidirectional nature of joint variability [15]. Thus, by construction, $C = C^T$, confirming the matrix's symmetric property [15].

**Positive Definite:** In addition to being symmetric, the covariance matrix must also be *positive definite* to ensure that it lies in the space of *SPD* matrices [15]. A matrix $C$ is considered positive definite if, for any non-zero vector $v \epsilon \mathbb{R}^n$, the following condition holds $v^T C v > 0$.

This property ensures the covariance matrix captures data variability so that no linear combination has zero or negative variance. The quadratic form $v^T C v$ represents the variance of the data projected along $v$, which must remain positive for real data. The covariance matrix is positive definite because its elements represent variances or covariances, which are non-negative. Since the variance of any non-trivial EEG signal is greater than zero, it follows that $v^T C v > 0$ for all non-zero $v$, confirming the matrix's positive definiteness. Thus, the covariance matrix in our EEG classification pipeline meets the symmetry and positive definiteness conditions, ensuring it lies within the Riemannian manifold of SPD matrices. This manifold underpins our feature engineering and classification processes, enabling the use of geometric properties for improved accuracy.

**C. Radial Basis Function (RBF) kernels**: The covariance matrix captures only linear relationships, so we use RBF kernels to capture non-linear ones. The RBF kernel maps data to a higher-dimensional space, improving the representation of non-linear relationships and boosting classification performance in BCI applications [16]. For two EEG signal samples $x_i \epsilon \mathbb{R}^n$ and $x_j \epsilon \mathbb{R}^n$. the RBF kernel function is defined as follows:

$$K(x_i, x_j) = \exp(-\frac{\|x_i - x_j\|^2}{2\sigma^2}) \quad (2)$$

where $\|x_i - x_j\|^2$ denotes the squared Euclidean distance between the two samples, and $\sigma$ is the bandwidth parameter that controls the spread of the kernel. Given a set of $N$ EEG signal samples $\{x_1, x_2, \ldots, x_N\}$, the RBF kernel matrix $K \in \mathbb{R}^{n \times n}$ is constructed such that:

$$K_{ij} = K(x_i, x_j) = \exp(-\frac{\|x_i - x_j\|^2}{2\sigma^2}) \quad (3)$$

This matrix captures the similarity between all pairs of EEG signals, effectively modeling their non-linear relationships.

**Symmetry:** The RBF kernel matrix $K$ is symmetric by construction. For any pair of indices $i$ and $j$, we have:

$$K_{ij} = \exp\left(-\frac{\|x_i - x_j\|^2}{2\sigma^2}\right) = K_{ji} \quad (4)$$

This equality holds because the Euclidean distance $\|x_i - x_j\|$ is symmetric, i.e. $\|x_i - x_j\| = \|x_j - x_i\|$. Therefore, the RBF kernel matrix satisfies the condition $K = K^T$, confirming that it is symmetric.

**Positive Definiteness of the RBF Kernel Matrix**: To establish that the RBF kernel matrix is positive definite, we must demonstrate that for any non-zero vector $v \epsilon \mathbb{R}^n$, the following holds $v^T K v > 0$. The quadratic form for the RBF kernel matrix is given by:

$$v^T K v = \sum_{i=1}^{N}\sum_{j=1}^{N} v_i v_j K_{ij} \quad (5)$$

Substituting the definition of the RBF kernel, we have:

$$v^T K v = \sum_{i=1}^{N}\sum_{j=1}^{N} v_i v_j \exp\left(-\frac{\|x_i - x_j\|^2}{2\sigma^2}\right) \quad (6)$$

The RBF kernel matrix is positive semi-definite because the exponential function is always positive for all $x_i$ and $x_j$, except when $v = 0$. It becomes strictly positive definite when samples are distinct. This makes the RBF kernel matrix symmetric and positive definite, making it suitable for use in the Riemannian manifold framework for EEG signal classification.

By ensuring that both the covariance matrices and the RBF kernel are symmetric and positive definite, we work within a Riemannian manifold, specifically the space of SPD matrices ($S_N^+$). Using these matrices within this framework enhances the robustness of our classification methods, providing a deeper understanding of the data's structure and improving machine learning performance for EEG signal classification.

*2) Phase 2: Data Clustering*

In this phase, we cluster the data points enriched by features extracted in the previous phase, now residing in the Riemannian manifold space. The goal is to preserve the local geometric context of the data, enabling *locally sensitive clustering-based classification* that retains the manifold's inherent structure. As described in Algorithm I, we propose a novel clustering algorithm designed for the manifold space, addressing the limitations of traditional K-Means, which uses Euclidean distance and fails to respect the non-Euclidean geometry of the manifold.

Our approach replaces Euclidean distance with a more suitable metric: *a weighted combination of Riemannian distance (measuring geodesic distance) and the slope of the tangent plane (providing local curvature information)*. The Riemannian distance ensures *global consistency*, while the tangent slope adds *local sensitivity*. The weights are selected through trial and error, normalized to sum to one, optimizing clustering performance. This method preserves the manifold's geometric structure and captures local curvature information, ensuring robust and locally sensitive classification in subsequent phases.

*3) Phase 3: Dimensionality Reduction*

At this stage, data points are clustered, preserving their local geometric information within the Riemannian manifold space. The next step in MPEC is dimensionality reduction. Here, we project all data points within each cluster onto the tangent space of the manifold, transferring them from the high-dimensional manifold space to a lower-dimensional Euclidean space. This simplifies the subsequent classification task. Dimensionality reduction improves classification performance by projecting data onto a tangent space, which retains the essential local structure while eliminating redundant features, reducing noise, and preventing overfitting [4]. The tangent space provides a linear approximation of the manifold, enabling the use of efficient machine learning techniques in a lower-dimensional space. This projection simplifies the data representation while preserving the intrinsic manifold structure, ensuring that locality and sensitivity are maintained. Ultimately, this step enhances the efficiency and accuracy of EEG classification while preserving the integrity of the original data's manifold context.

*4) Phase 4: Ensemble Modelling*

After feature engineering, clustering, and dimensionality reduction, we've created a strong foundation for accurate EEG classification by preserving the Riemannian manifold structure. This approach maintains local geometric relationships in the data, enhancing accuracy. To further improve performance, we propose ensemble modeling, combining multiple weak learners to reduce variance, minimize overfitting, and improve generalization. The main ensemble strategies are *bagging*, which reduces variance by averaging predictions from random data subsets; *boosting*, which sequentially corrects errors from previous learners; and *stacking*, which uses parallel models combined by a meta-model for the final prediction, each offering different ways to enhance model stability and accuracy [17] [18]. Ensemble learning improves performance by combining multiple models.

We chose stacking for our ensemble method because it integrates diverse learners more effectively than bagging and boosting [17]. Stacking uses a meta-model, to combine base learner predictions, utilizing the preserved manifold structure for better adaptability.

After feature engineering and dimensionality reduction, we optimized EEG classification by preserving the Riemannian manifold structure. For stacking, we used a *Ridge Classifier* as the meta-model, leveraging its efficiency in handling high-dimensional data and its L2 regularization to minimize overfitting, ensuring robustness against noise and multicollinearity (key for the complex patterns and correlations in EEG data). By using it, we effectively combine base learner predictions while maintaining the integrity of the preserved manifold structure, resulting in a balanced and accurate classification performance. We use four distinct weak learners in the stacking ensemble, each chosen to capture different data aspects, ensuring diversity and enhancing classification robustness while complementing our manifold-preserving approach:

**A. Support Vector Machine (SVM)**: SVMs are powerful models for classification and regression, especially with high-dimensional data like EEG, as they find the best hyperplane to separate classes and avoid overfitting. Their ability to capture complex patterns makes them useful in our ensemble, improving the detection of subtle brain activity differences and boosting classification performance.

**B. Logistic Regression (LR):** LR is a statistical method that models the relationship between input features and class probabilities using a logistic function, offering an intuitive interpretation of feature effects. It is included as a weak learner, it is efficient, interpretable, and captures linear patterns, helping stabilize the ensemble and balance complex classifiers in EEG classification.

**C. Multi-Layer Perceptron (MLP)**: MLPs are neural networks with multiple layers that can capture non-linear relationships in data, making them effective for complex tasks like EEG classification. Their ability to model intricate patterns adds valuable diversity to the ensemble, improving the final decision.

**D. Random Forest**: it builds multiple decision trees from random data subsets and averages their predictions, reducing overfitting and improving robustness. As a weak learner, it handles EEG data noise well and boosts performance by increasing diversity among ensemble models.

This *ensemble of clustering-based classifiers* preserves and utilizes manifold information for more accurate EEG classification. By combining the local geometric properties maintained throughout the clustering and dimensionality reduction phases with the strengths of ensemble learning, our approach significantly enhances classification performance. This ensemble strategy fully utilizes manifold-related structure and local information, creating a classifier that is both robust and sensitive to EEG signal patterns. Algorithm II provides the pseudocode of our methodology.

## IV. RESULTS

*A. Evaluation Metrics*

In our evaluations, we employed the following standard classification metrics: *precision*, *recall*, *F1-measure*, and

| **Algorithm I**: The Proposed K-Means Clustering Algorithm in Riemannian Manifold Space |
|---|
| **1. Initialize:**<br>　1.1. Select $k$ initial cluster centroids in the manifold space.<br>　1.2. Initialize the weight parameters $w_1$ and $w_2$, where $w_1 + w_2 = 1$.<br>**2. Assign Points to Clusters:**<br>　2.1. For each data point $x_i$:<br>　　2.1.1. Compute the Riemannian distance $d_R(x_i, c_j)$ between the point $x_i$ and each centroid $c_j$.<br>　　2.1.2. Calculate the slope of the tangent plane at $x_i$ in relation to each centroid $\theta_{x_i}$.<br>　　2.1.3. Normalize the $d_R(x_i, c_j)$ and $\theta_{x_i}$ in a range between 0 and $1$.<br>　　2.1.3. Combine these two metrics using the weighted sum<br>$$D(x_i, c_j) = w_1 d_R(x_i, c_j) + w_1 \theta_{x_i}$$<br>　　2.1.4. Assign $x_i$ to the cluster with the minimum combined distance $D(x_i, c_j)$.<br>**3. Update Cluster Centroids:**<br>　3.1. For each cluster, update the centroid by calculating the mean of all points assigned to the cluster in the manifold space.<br>**4. Repeat:**<br>　4.1. Repeat steps 2 and 3 until convergence (i.e., cluster assignments no longer change or the change is below a predefined threshold).<br>**5. Output:**<br>　5.1. The final set of cluster centroids and the cluster assignments of all points. |

| **Algorithm II**: Step-by-Step Description of the Proposed Method (MPEC) |
|---|
| **Phase 1: Feature Engineering**<br>　1.1 Input the raw EEG signals from 22 channels.<br>　1.2 Compute the covariance matrix for each EEG signal to capture the linear relationships between the channels.<br>　1.3 Apply the RBF kernel to extract non-linear relationships among the EEG channels.<br>　1.4 Combine the covariance matrix and RBF kernel matrices using a weighted sum, with weights incremented by 0.1, where the sum of weights is 1. These weights are selected through trial and error for optimal performance.<br>**Phase 2: Data Clustering**<br>　2.1 Localize the data points (i.e., feature-enriched EEG signals) within the Riemannian manifold space.<br>　2.2 Propose a novel clustering algorithm that relies on a modified K-Means, but with a new distance metric suitable for manifold space:<br>　　2.2.1 Calculate the Riemannian distance between data points and centroids.<br>　　2.2.2 Measure the slope of the tangent plane at each point.<br>　　2.2.3 Combine these values using a weighted sum for clustering.<br>　2.3 Assign each data point to its closest cluster based on the combined distance metric.<br>　2.4 Update cluster centroids and repeat until convergence.<br>**Phase 3: Dimensionality Reduction**<br>　3.1 Project each data point within a cluster onto the corresponding tangent space of the manifold.<br>　3.2 Reduce the dimensionality by mapping the high-dimensional manifold data into a lower-dimensional Euclidean space while preserving the local geometric structure.<br>　3.3 Maintain the manifold-related information in the lower-dimensional space to improve classification performance.<br>**Phase 4: Ensemble Modeling**<br>　4.1 Perform classification by applying an ensemble learning approach (*stacking*)<br>　　4.1.1 Use four weak learners to capture different aspects of the data, including: *SVM*, *LR*, *MLP*, and *Random Forest*.<br>　　4.1.2 Each weak learner is trained on the data clusters from the reduced-dimensional space.<br>　4.2 Combine the outputs of the weak learners using a simple *Ridge Classifier* as the meta-model.<br>　4.3 Make the final classification decision by stacking the weak learners' predictions through the meta-model.<br>　4.4 Output the classified EEG signals with preserved manifold structure. |

*accuracy*, with a particular emphasis on *accuracy* as the primary evaluation metric.

### B. Parameter Setting

This section outlines the parameter settings for each phase of the MPEC EEG classification system, covering *feature engineering*, *clustering*, *dimensionality reduction*, and *ensemble modeling*. These settings are key to optimizing performance and ensuring a consistent processing pipeline. The train-test split is set at 80-20 for all models.

**Phase 1-Feature engineering**: in this step, we analyzed the correlation between EEG signals and target labels to select relevant features, testing subsets of 10 to 22 features. The best performance was achieved with 22 features, using the full covariance matrix with values normalized between 0 and 1. The RBF kernel's width parameter (*sigma*) was set to 0.1 to better capture non-linear relationships and improve accuracy.

**Phase 2-Data clustering:** in this phase, we used our new K-Means method to preserve the local structure of EEG data, testing different cluster numbers (K) from 3 to 13. The best results came with $K = 3$, which captured the data's core structure most effectively.

**Phase 3-Dimensionality reduction:** no additional parameter settings were required.

**Phase 4-Ensemble modelling:** We used four weak learners (*SVM*, *LR*, *MLP*, and *random forest*) in our ensemble model, with parameters details including: the *SVM model* utilizes the RBF as its kernel, which plays a critical role in defining the decision boundary for classification. Additionally, L1 regularization is applied to minimize the model's generalization error. The *LR model* is configured with the solver (*lbfgs*) set to a maximum of 1,000 iterations to ensure convergence. Additionally, L1 regularization is applied to reduce the model's generalization error. The *Random Forest model* is configured with the parameter *n_estimators* set to 100, which specifies the number of decision trees in the ensemble. The maximum depth of the trees is set to *None*, meaning that nodes are expanded until all leaves are pure or contain fewer samples than the minimum required to split an internal node. Table I demonstrates the parameter configuration of the proposed MLP model.

The Ridge Classifier meta-learner in our stacking ensemble was configured with regularization strength parameter, (*alpha*) was set to *10*, (striking a balance between bias and variance). The *solver* was set to "*auto*," allowing the algorithm to automatically select the most suitable optimization method based on the data. The tolerance for convergence was set to $1 \times 10^{-4}$, ensuring precise optimization while maintaining computational efficiency. Additionally, a *5-fold* cross-validation was employed to validate the model and fine-tune its performance, minimizing overfitting and ensuring reliable predictions.

TABLE I. THE PARAMETER SETTINGS OF THE PROPOSED MLP MODEL

| *Parameter* | *Setting* |
|---|---|
| Number of epochs | 100 |
| Activation | Tanh |
| Alpha | 0.0001 |
| Hidden layer sizes | 100 |
| Learning rate | 0.001 |
| Optimizer | Adam |

TABLE II. EVALUATION RESULTS (IN PERCENT) FOR PROPOSED EEG CLASSIFICATION MODELS

|  | **Precision** | **Recall** | **F1-Score** | **Accuracy** |
|---|---|---|---|---|
| **SVM** | 76.31 | 74.52 | 75.90 | 75.21 |
| **LR** | 77.74 | 75.59 | 76.65 | 76.56 |
| **MLP** | 75.90 | 73.63 | 74.75 | 74.63 |
| **Random Forest** | 71.58 | 69.39 | 70.47 | 70.39 |
| **MPEC (ensemble)** | **78.16** | **76.42** | **77.28** | **78.12** |

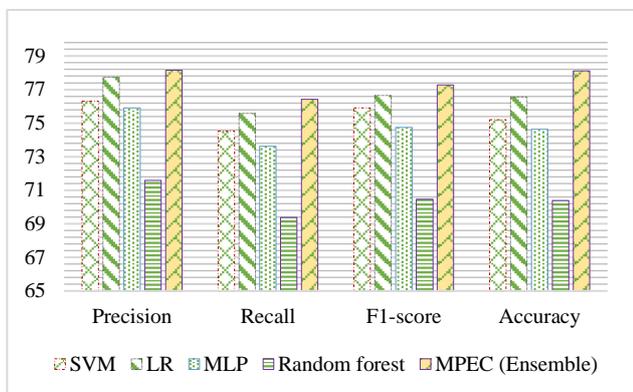

Fig. 2. Visual representation of evaluation results (in percent) for proposed EEG classification models

### C. Evaluation Results

Table II presents the average evaluation results of four EEG classification models and MPEC, our primary ensemble method, for all nine subjects in the dataset. It includes precision, recall, F1-measure, and accuracy. Since the F1-measure overlooks true negatives, it may not fully reflect system reliability in practice [19][20]. Thus, accuracy is emphasized as the key metric for model credibility. Fig. 2 provides a visual comparison of the performance of all EEG classification models proposed in this study.

### D. Baseline Models

To evaluate the performance of our proposed EEG classification system, MPEC, we compared it with baseline models that utilize the BCI Competition IV Dataset 2a in multi-label classification.

■ **Jindal et al.** [21]: This study proposes a new method combining MDA-SOGWO for channel selection and CCA-RFE for feature optimization in multi-class MI-BCI, effectively enhancing classification for BCI system design.

■ **Hou et al.** [22]: The paper presents a novel MI-BCI classification framework using Bispectrum, Entropy, and Common Spatial Pattern (BECSP) for feature extraction, achieving optimal results with SVM and an RBF kernel.

■ **Wijaya et al.** [23]: the paper introduces a two-stage detection and voting scheme with a one-versus-one approach to enhance multi-class EEG motor imagery classification in BCI systems.

■ **Rodrigues et al.** [24]: this study evaluates frameworks for classifying motor imagery in EEG-based BCIs, finding that a recurrence-based approach best captures nonlinear relations, while eigenvector centrality is most efficient for processing time.

■ **Amin et al.** [25]: this study presents a multi-layer CNN fusion approach for EEG motor imagery classification, combining diverse CNN architectures to capture spatial and temporal features.

### E. Discussion

The MPEC (*Manifold-Preserved EEG Classification via an Ensemble of Clustering-Based Classifiers*) has shown better accuracy in classifying EEG signals from nine subjects in the BCI Competition IV dataset 2a compared to the baseline models, as detailed in Table III and Fig. 3. By preserving the manifold structure of EEG data, MPEC captures non-linear, high-dimensional relationships among signals, outperforming traditional baseline models that use Euclidean distances. This approach addresses the limitations of methods that distort geometric relationships, reducing classification performance. MPEC's robustness comes from its ensemble, integrating various classifiers that offer unique insights into EEG patterns, allowing it to handle the complex and variable nature of brain signals. Let's to answer the research questions stated before in Introduction:

**RQ1**: Preserving the manifold structure has indeed enhanced classification accuracy. The 78.12% accuracy across all subjects (Table III and Fig. 3) indicates a substantial improvement over conventional Euclidean-based models, validating the effectiveness of manifold preservation in achieving high-performance classification.

**RQ2**: the ensemble classification on locally clustered Riemannian manifolds in MPEC improves performance by combining weak learners in a manifold-aware framework, enhancing sensitivity to EEG signal variations compared to traditional, approaches.

**RQ3**: the novel K-means algorithm, using Riemannian distance and curvature, effectively clusters high-dimensional EEG data by preserving its geometry, reducing noise and variance for more accurate classification.

**RQ4**: projecting data points from Riemannian clusters onto lower-dimensional tangent spaces preserves discriminative power, maintaining strong classification accuracy while retaining essential geometric information for efficient processing.

## V. CONCLUSION

This study addressed the challenge of accurately classifying EEG signals while preserving the non-Euclidean, manifold structure inherent in EEG data. To overcome the limitations of traditional methods that disregard manifold geometry, we proposed the *Manifold-Preserved EEG Classification via an Ensemble of Clustering-Based Classifiers (MPEC)*. MPEC integrates four weak learners (SVM, Logistic Regression, random forest, and MLP) with *Ridge Classifier* as the meta-model in a stacking-based ensemble. This ensemble approach effectively combines the strengths of individual learners, resulting in a more robust and accurate classification model. Guided by our main research question, MPEC successfully meets its objective by preserving the manifold structure, thus maintaining essential local geometric relationships within the data. The proposed method provides a high-performance EEG classification system that leverages manifold context, advancing the accuracy and reliability of EEG analysis in BCIs and related applications.

TABLE III. COMPARISON OF BASELINE MODEL ACCURACIES FOR EEG SIGNAL CLASSIFICATION ON THE BCI COMPETITION IV DATASET 2A AGAINST THE PROPOSED MODEL (MPEC) FOR ALL 9 SUBJECTS IN THE DATASET.

| | Subj. 1 | Subj. 2 | Subj. 3 | Subj. 4 | Subj. 5 | Subj. 6 | Subj. 7 | Subj. 8 | Subj. 9 | Avg. |
|---|---|---|---|---|---|---|---|---|---|---|
| **Jindal et al.** [21] | 60.91 | 50.57 | 42.52 | 78.16 | 67.81 | 52.87 | 65.51 | **94.25** | **90.80** | 67.04 |
| **Hou et al.** [22] | 80.90 | 64.24 | 85.76 | 65.63 | 44.79 | 54.17 | 84.38 | 84.03 | 80.56 | 71.61 |
| **Wijaya et al.** [23] | 77.43 | **82.99** | 76.39 | 65.28 | **73.26** | **69.79** | 52.78 | 82.99 | 81.25 | 73.57 |
| **Rodrigues et al.** [24] | 60.00 | 33.00 | 67.00 | 45.00 | 33.00 | 33.00 | 35.00 | 70.00 | 67.00 | 49.00 |
| **Amin et al.** [25] | 62.07 | 42.44 | 63.12 | 52.09 | 49.96 | 37.16 | 62.54 | 59.32 | 69.43 | 55.34 |
| **MPEC** (proposed method) | **87.68** | 67.72 | **90.73** | 76.92 | 55.14 | 65.92 | **88.76** | 90.74 | 79.46 | **78.12** |

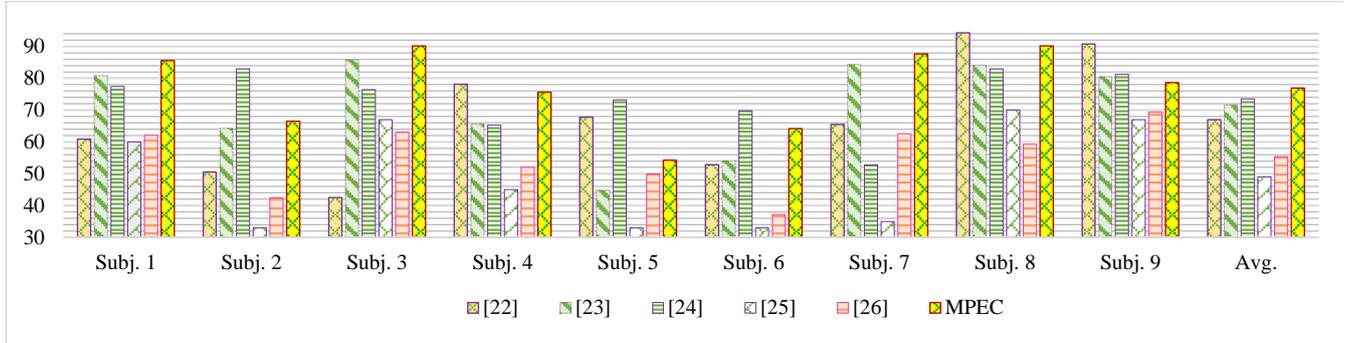

Fig 3. Visual comparison of baseline model accuracies for EEG signal classification on the BCI Competition IV dataset 2a against the proposed model (MPEC) for all 9 subjects in the dataset.